\documentclass[wcp]{jmlr}

\usepackage{longtable}

\usepackage{booktabs}
\usepackage{graphicx} 
\usepackage{natbib}
\usepackage{algorithmic}
\usepackage{hyperref}
\usepackage{comment}
\usepackage{amsmath}
\usepackage{multirow}
\usepackage{amssymb}
\usepackage{stmaryrd}
\usepackage{bm}

\newcommand{\KK}{K}

\newcommand{\sets}[1]{\{#1\}}
\newcommand{\ind}[1]{{\boldsymbol{1}}_{\sets{#1}}}

\newcommand{\Fcal}{\mathcal{F}} 
\newcommand{\Gcal}{\mathcal{G}}
\newcommand{\Xcal}{\mathcal{X}} 
\newcommand{\Ycal}{\mathcal{Y}} 

\newcommand{\Mcal}{\mathcal{M}}

\newcommand{\var}{\textbf{Var}}
\newcommand{\cov}{\textbf{Cov}}
\newcommand{\argmax}{\textbf{argmax}}

\newcommand{\maximize}{\textbf{max}}
\newcommand{\yb}{{\bf y}}

\newcommand{\gb}{{\bf g}}

\newcommand{\wb}{{\bf w}}

\newcommand{\ub}{{\bf u}}

\newcommand{\mub}{\mathbf{\mu}}

\newcommand{\vell}{\mathbf{\ell}}

\newcommand{\ip}[2]{\langle #1, #2 \rangle}

\jmlrvolume{29}
\jmlryear{2013}
\jmlrworkshop{ACML 2013}

\title[Random Graph Ensembles]{Multilabel Classification through Random Graph Ensembles}

  \author{\Name{Hongyu Su} \Email{hongyu.su@aalto.fi} \\
       \Name{Juho Rousu} \Email{juho.rousu@aalto.fi} \\
       \addr Helsinki Institute of Information Technology (HIIT)\\
        Department of Information and Computer Science\\
       Aalto University, Konemiehentie 2, 02150 Espoo, Finland
 }

\editor{Cheng Soon Ong and Tu Bao Ho}

\begin{document}

\maketitle

\begin{abstract}

We present new methods for multilabel classification, relying on ensemble learning on a collection of random output graphs imposed on the multilabel and a kernel-based structured output learner as the base classifier. For ensemble learning, differences among the output graphs provide the required base classifier diversity and lead to improved performance in the increasing size of the ensemble. We study different methods of forming the ensemble prediction, including majority voting and two methods that perform inferences over the graph structures before or after combining the base models into the ensemble. We compare the methods against the state-of-the-art machine learning approaches on a set of heterogeneous multilabel benchmark problems, including multilabel AdaBoost, convex multitask feature learning, as well as single target learning approaches represented by Bagging and SVM. In our experiments, the random graph ensembles are very competitive and robust, ranking first or second on most of the datasets. Overall, our results show that random graph ensembles are viable alternatives to flat multilabel and multitask learners.

\end{abstract}

\begin{keywords}
multilabel classification; structured output; ensemble methods; kernel methods; graphical models
\end{keywords}

%
%
\section{Introduction}

Multilabel and multitask classification rely on representations and learning methods that allow us to leverage the dependencies between the different labels.
When such dependencies are given in form of a graph structure such as a sequence, a hierarchy or a network, structured output prediction  \citep{TGK:nips03,THJA:icml04,rousu2006kbl} becomes a viable option, and has achieved a remarkable success.
For multilabel classification, limiting the applicability of the structured output prediction methods is the very fact they require the predefined output structure to be at hand, or alternatively auxiliary data where the structure can be learned from.
When these are not available, flat multilabel learners or collections of single target classifiers are thus often resorted to.

In this paper, we study a different approach, namely using ensembles of graph labeling classifiers, trained on randomly generated {output graph} structures.  
The methods are based on the idea that variation in the graph structure shifts the inductive bias of the base learners and causes diversity in the predicted multilabels. 
Each base learner, on the other hand, is trained to predict as good as possible multilabels, which make them satisfy the weak learning assumption, necessary for successful ensemble learning. 
 
%

Ensembles of multitask or multilabel classifiers have been proposed, but with important differences.
The first group of methods, boosting type, rely on changing the weights of the training instances so that difficult to classify instances gradually receive more and more weights.
The AdaBoost boosting framework has spawned multilabel variants  \citep{SchSin00,Esu08}.
In these methods the multilabel is considered essentially as a flat vector.
The second group of methods, Bagging, are based on bootstrap sampling the training set several times and building the base classifiers from the bootstrap samples.
Thirdly, randomization has been used as the means of achieving diversity by \citet{yan2007model} who select different random subsets of input features and examples to induce the base classifiers, and by  \citet{su2011multi} who use majority voting over random graphs in drug bioactivity prediction context.
Here we extend the last approach to two other types of ensembles and a wider set of applications, with gain in prediction performances.

The remainder of the article is structured as follows.
In section \ref{sec_graph_labeling} we present the structured output model used as the graph labeling base classifier.
In Section \ref{sec_ensembles} we present three multilabel ensemble learning methods based on the random graph labeling.
In section \ref{sec_experiments} we present empirical evaluation of the methods. 
In section \ref{sec_conclusions} we present concluding remarks. 

%
%

\section{Multilabel classification through graph labeling}
\label{sec_graph_labeling}

We examine the following multilabel classification setting.
We assume data from a domain $\Xcal \times \Ycal$,
where $\Xcal$ is a set and $\Ycal = \Ycal_1 \times \dots \times \Ycal_k$ is the set of multilabels, represented by a Cartesian product of the sets $\Ycal_j = \sets{1,\dots,l_j}, j = 1,\dots,k$.
A vector $\yb = (y_1,\dots,y_k) \in \Ycal$ is called the {\em multilabel} and the components $y_j$ are called the {\em microlabels}. 
We assume that a training set $\lbrace{(x_i,\yb_i)\rbrace}_{i=1}^m \subset \Xcal \times \Ycal$ has been given. 
A pair $(x_i,\yb)$ where $x_i$ is a training pattern and $\yb \in \Ycal$ is arbitrary, is called a {\em pseudo-example},
to denote the fact that the pair may or may not be generated by the distribution generating the training examples. The goal is to learn a model $F: \Xcal \mapsto \Ycal$ so that the expected loss 
over predictions on future instances is minimized, where the loss is  chosen suitably for multilabel learning problems. By $\ind{\cdot}$ we denote the indicator function $\ind{A} = 1$, if  $A$ is true, $\ind{A} = 0$ otherwise.

Here, we consider solving multilabel classification with graph labeling classifiers that, in addition to the training set, assume a graph $G = (V,E)$ with nodes $V = \{1,\dots,k\}$ corresponding to
microlabels and edges $E \subset V \times V$ denoting potential dependencies between the microlabels. For an edge $e = (j,j') \in E$, by  $\yb_e = (y_j,y_{j'})$   we denote the {edge label} of $e$ in multilabel $\yb$, induced by concatenating the microlabels corresponding to end points of $e$, with corresponding domain of edge labelings $\Ycal_e = \Ycal_j \times \Ycal_{j'}$. By $\yb_{ie}$ we denote the label of the edge $e$ in the $i$'th training example. We also denote by $u_j$ the possible label of node $j$, and by $\ub_e$ the possible label of edge $e$. Naturally, $u_j\in \Ycal_j$ and $\ub_e\in \Ycal_e$. See supplementary material for a complete list of notations.

\subsection{Graph labeling classifier}

As the graph labeling classifier in this work we use max-margin structured prediction, which aims to learn a
compatibility score  
\begin{align}
\psi(x,\yb) = \ip{w}{\varphi(x,\yb)} = \sum_{e \in E} \ip{w_{e}}{\varphi_{e}(x,\yb_e)}= \sum_{e \in E} \psi_e(x,\yb_e)
\label{compatibility_score}
\end{align}
between an input $x$ and a multilabel $\yb$, where  by $\ip{\cdot}{\cdot}$ we denote the inner product and $\psi_e(x,\yb_e)$ 
is a shorthand for the compatibility score, or potential, between an edge label $\yb_e$ and the object $x$.
The joint feature map 
  $$\varphi(x,\yb) = \phi(x)\otimes \Upsilon(\yb) =  \phi(x)\otimes\left(\Upsilon_e(\yb_e)\right)_{e \in E} = \left(\varphi_e(x,\yb_e)\right)_{e \in E}$$
is given by a tensor product of an input feature $\phi(x)$ and the feature space embedding of the multilabel $\Upsilon(\yb) =  \left(\Upsilon_e(\yb_e)\right)_{e \in E}$, consisting of edge labeling indicators  
$\Upsilon_e(\yb_e) = \left(\ind{\yb_e = \ub_e}\right)_{\ub_e \in \Ycal_e}$. 
The benefit of the tensor product representation is that context (edge labeling) sensitive weights can be learned for input features and no prior alignment of input and output features 
needs to be assumed.


The parameters of the model are learned through max-margin optimization, where
the primal optimization problem takes the form \citep[e.g.][]{TGK:nips03,THJA:icml04,rousu2006kbl}
\begin{align}
\underset{w}{{\mathbf{min }} } &\ \frac{1}{2}||{w}||^2 + C \sum_{i=1}^m \xi_i
\label{soft_margin}\\
\text{ s.t. } & \ip{w}{\varphi(x_i,\yb_i)} \ge \underset{\yb \in \Ycal}{\argmax} \left(\ip{w}{\varphi(x_i,\yb)} + \ell(\yb_i,\yb)\right) - \xi_i, \nonumber \\ 
& \text{ for  } i=1,\dots,m \nonumber
\end{align}
where $w$ contains the weights to be learned,  $\xi_i$ denotes the slack allotted to each example, $\ell(\yb_i,\yb)$ is the loss between pseudo-labeling and correct labeling and $C$ is the slack parameter that controls the amount of regularization in the model.  The primal form can be interpreted as maximizing the loss-scaled margin between the correct training example and incorrect pseudo-examples. 
The Lagrangian dual form of (\ref{soft_margin}) is given as
\begin{align}
{\underset{\alpha \geq 0}{ \mathbf{\maximize } }}\ & \alpha^T \vell - \frac{1}{2}\alpha^T \KK \alpha  \label{orig_dual} \\
  \text{ s.t.}  & \sum_{\yb} \alpha(i,\yb) \le C, \forall i=1,\dots,m \text{ and } \yb \in \Ycal, \nonumber
\end{align}
where $\alpha = \left(\alpha(i,\yb)\right)_{i,\yb}$ denotes the dual variables and $\vell =\left(\ell(\yb_i,\yb)\right)_{i,\yb}$ the loss for each pseudo-example $(x_i,\yb)$.
The joint kernel
\begin{align*}
  \KK(x_i,\yb;x_j,\yb') &= \langle \varphi(x_i,\yb_i) - \varphi(x_i,\yb),  \varphi(x_j,\yb_j) - \varphi(x_j,\yb')\rangle \\
    & = \langle \phi(x_i),\phi(x_j)\rangle_\phi \cdot \langle(\Upsilon(\yb_i)-\Upsilon(\yb),\Upsilon(\yb_j)-\Upsilon(\yb')\rangle_\Upsilon \\
    & = K_\phi(x_i,x_j) \cdot \left(K_\Upsilon(\yb_i,\yb_j)-K_\Upsilon(\yb_i,\yb')-K_\Upsilon(\yb,\yb_j)+K_\Upsilon(\yb,\yb')\right) 
\end{align*}
is composed by product of input $K_\phi(x_i,x_j) = \langle x_i,x_j \rangle_\phi$ and output $K_\Upsilon(\yb,\yb') =  \ip{\yb'}{\yb}_\Upsilon = \sum_e K_{\Upsilon,e}(\yb_e,\yb'_e)$ kernels,
with $K_{\Upsilon,e}(u,u') = \ip{\Upsilon_e(u)}{\Upsilon_e(u')}_\Upsilon$. 


\subsection{Factorized dual form}

The model (\ref{orig_dual}) is transformed to the factorized dual form, where the edge-marginals of dual variables are used in place of the original dual variables
\begin{equation}
  \mu(i,e,\ub_e) = \sum_{\yb \in \Ycal} \ind{\Upsilon_e(\yb) = \ub_e}\alpha(i,\yb), \label{eq_marginal}
\end{equation}
where $e  = (j,j') \in E$ is an edge in the output network and $\ub_e \in \Ycal_j \times \Ycal_{j'}$ 
is a possible labeling for the edge $(j,j')$.
Using the factorized dual representation, we can state the dual problem (\ref{orig_dual}) in equivalent form \citep[c.f.][]{TGK:nips03,rousu2007} as 
\begin{equation}{\underset{\mu \in \Mcal}{ \max }}\   \mub^T\vell - 
  \frac{1}{2} \mub^T K_{\Mcal} \mub, \label{marg_dual}
\end{equation}
where $\vell = \left(\ind{\yb_{ie} \neq \ub_e})\right)_{i,e,\ub_e}$ is the vector of losses between the edge-labelings, and $\mub = \left(\mu(i,e,\ub_e)\right)_{i,e,\ub_e} \in \Mcal$ is the vector of marginal dual variables lying in 
the marginal polytope \citep[c.f.][]{wainwright2005map}
$$\Mcal = \{\mu | \exists \alpha \text{ s.t. } \mu(i,e,\ub_e) = \sum_{\yb \in \Ycal} \ind{\yb_{ie} = \ub_e} \alpha(i,\yb), \forall i,\ub_e,e\}$$
of the dual variables,  the set of all combinations of marginal dual variables (\ref{eq_marginal}) of 
a training examples that correspond to some $\alpha$ in the original dual feasible set in (\ref{orig_dual}). The factorized joint kernel is given by $K_\Mcal = diag(K_{\varphi,e})_{e \in E}$,
where 
\begin{multline*}
   K_{\varphi,e}(x_i,\yb_e;x_j,\yb_e')    \\ 
         = K_\phi(x_i,x_j) \cdot \left(K_{\Upsilon,e}(\yb_{ie},\yb_{je})-K_{\Upsilon,e}(\yb_{ie},\yb_e)-K_{\Upsilon,e}(\yb_e,\yb_{je})+K_{\Upsilon,e}(\yb_e,\yb_e')\right)
\end{multline*}
containing the joint kernel values pertaining to the edge $e$. 

The factorized dual problem (\ref{marg_dual}) is a quadratic program with a number of variables {\em linear} in both the size of the output network and the number of training examples.
There is an exponential reduction in the number of dual variables from the original dual (\ref{orig_dual}), however, with the penalty of more complex feasible 
polytope. For solving (\ref{marg_dual}) we use MMCRF  \citep{rousu2007} that relies on a  conditional gradient method. Update directions are found in linear time via probabilistic inference,  making use of the  the exact 
correspondence of  maximum margin violating multilabel in the primal (\ref{soft_margin}) and steepest feasible gradient of the dual objective (\ref{orig_dual}).

\subsection{Inference}

With the factorized dual, the compatibility score of labeling an edge $e$ as $\yb_e$ given input $x$ can be expressed in terms of kernels and marginal dual variables
as shown by the following lemma.
  
\begin{lemma}
\label{lemma_equ}
Let $w$ be the solution to (\ref{soft_margin}), $\varphi(x,\yb)$ be the joint feature map, and let $G=(E,V)$ be the graph defining the output graph structure, and let us denote
$$H_e(i,\ub_e;x,\yb_e)  = K_\phi(x,x_i)\cdot\left(K_{\Upsilon,e}(y_{ie},\yb_e)-K_{\Upsilon,e}(\ub_e,\yb_e)\right).$$
Then, we have
\begin{align*}
 \psi_e(x,\yb_e) =  \ip{w_e}{\varphi_e(x,\yb_e)}  =  \sum_{i,\ub_e} \mu(i,e,\ub_e) \cdot H_e(i,\ub_e;x,\yb_e),
\end{align*}
where $\mu$ is the marginal dual variable learned by solving optimization problem (\ref{marg_dual}).
\end{lemma}

\begin{proof}
See supplementary material.
\end{proof}

Consequently, the inference problem can be solved in the  factorized dual by
\begin{align}
  \hat\yb(x) = & \argmax_{\yb \in \Ycal}\sum_e \psi_e(x,\yb_e)  = \argmax_{\yb \in \Ycal} \sum_e  \langle \wb_e,\varphi_e(x,\yb_e) \rangle  \label{eq_inference} \\
  =  & \argmax_{\yb \in \Ycal} \sum_{e,i,\ub_e}  \mu(i,e,\ub_e) H_e(i,\ub_e;x,\yb_e) . \notag
  \end{align}
The inference problem (\ref{eq_inference}) is used not only in prediction phase to output multilabel $\hat\yb$ that is compatible with input $x$, but also in model training to find the pseudo-example $\yb$ that violates margin maximally. 
To solve (\ref{eq_inference}), any commonly used inference technique can be used.
In this paper we use MMCRF that relies on the message-passing method, also referred as loopy belief propagation (LBP).
We use early stopping in inference of LBP, so that the number of iterations is limited by the diameter of the output graph $G$.

%
%
\section{Learning graph labeling ensembles}
\label{sec_ensembles}

\begin{algorithm}[b]
\caption{Graph Labeling Ensemble Learning}
\label{alg_gle}
\begin{algorithmic}[1]
\REQUIRE Training sample $S = \lbrace (x_i,\yb_i)\rbrace_{i=1}^m$,
ensemble size $T$,
graph generating oracle function $outputGraph: t\in\{1,\dots,T\} \mapsto \Gcal_{k}$, aggregation function $A(\cdot): \Fcal \times \cdots \times \Fcal \mapsto \Ycal$
 
\ENSURE Multilabel classification ensemble $F(\cdot): \Xcal \mapsto \Ycal$
\FOR{$t \in \{1,\dots,T\}$}
\STATE $G^{(t)} = \text{outputGraph}(t)$
\STATE $F^{t}(\cdot) = \text{learnGraphLabelingClassifier}(\left(x_i\right)_{i=1}^{m},\left(\yb_i\right)_{i=1}^m,G^{(t)})$
\ENDFOR
\STATE $F(\cdot) = A(F^{(1)}(\cdot),\dots,F^{(T)}(\cdot))$
\end{algorithmic}
\end{algorithm}

In this section we consider generating ensembles of multilabel classifiers, where each base model is a graph labeling classifier.
Algorithm \ref{alg_gle} depicts the general form of the learning approach. We assume a function to output a random graph $G^{(t)}$ for each stage of the ensemble, 
a base learner to learn the graph labeling model $F^{(t)}(\cdot)$, and an aggregation function $A(\cdot)$ to compose the ensemble model.
The prediction of the model is then obtained by aggregating the base model predictions
\begin{align*}
	F(x) = A(F^{(1)}(x),\dots,F^{(T)}(x)).
\end{align*}

Given a set of base models trained on different graph structures we expect the predicted labels of the ensemble have diversity which is known to be necessary for ensemble learning.
At the same time, since the graph labeling classifiers aim to learn accurate multilabels, we expect the individual base classifiers to be reasonably accurate, irrespective of the 
slight changes in the underlying graphs. Indeed, in this work we use randomly generated graphs to emphasize this point.
We consider the following three aggregation methods: 
\begin{itemize} 
	\item  In {\em majority voting ensemble}, each base learner gives a prediction of the multilabel. The ensemble prediction is obtained by taking the most frequent value for each microlabel. 
Majority voting aggregation is admissible for any multilabel classifier.
\end{itemize}
Second, we consider two aggregation strategies that assume the base classifier has a conditional random field structure:  
\begin{itemize}  
	\item In {\em average-of-maximum-marginals aggregation}, each base learner infers local maximum marginal scores for each microlabel. The ensemble prediction is taken as the value with highest average local score.   
	\item In {\em maximum-of-average-marginals aggregation}, the local edge potentials of each base model are first averaged over the ensemble and maximum global marginal scores are inferred from the averages. 
\end{itemize}
In the following, we detail the above aggregation strategies.

\subsection{Majority voting ensemble (MVE)}


The first ensemble model we consider is the majority voting ensemble (MVE), 
which was introduced in drug prediction context by \citet{su2011multi}.
In MVE, the ensemble prediction or each microlabel is the most frequently appearing prediction among the base classifiers 
\begin{align*}
F_{j}^{\text{\tiny MVE}}(x) = \argmax_{y_j \in \Ycal_j} \left(\frac{1}{T} \sum_{i=1}^T \ind{F_{j}^{(t)}(x)=y_j}\right),
\end{align*}
where $F^{(t)}(x) = (F_j^{(t)}(x))_{j=1}^k$ is the predicted multilabel in $t$'th base classifier.
When using (\ref{marg_dual}) as the base classifier, predictions $F^{(t)}(x)$ are obtained via solving the inference problem (\ref{eq_inference}).
We note, however, in principle, any multilabel learner will fit into the MVE framework as long as it adapts to a collection of output graphs $\mathcal{G}=\{G^{(1)},\cdots,G^{(T)}\}$ and generates multilabel predictions accordingly from each graph.

%

%

\subsection{Average of Max-Marginal Aggregation (AMM)}

Next, we consider an ensemble model where we perform inference over the graph to extract information on the learned compatibility scores in each base models. Thus, 
we assume that we have access to the  compatibility scores between the inputs and edge labelings 
\begin{align*}
\Psi_E^{(t)}(x) = (\psi^{(t)}_{e}(x,\ub_e))_{e \in E^{(t)},\ub_e \in \Ycal_e}.
\end{align*}
In the Average of Max-Marginals (AMM) model, our goal is to infer for each microlabel $u$ of each node $j$
 its {\em max-marginal} \citep{wainwright2005map}, that is, the maximum score of a multilabel that is consistent with $y_j = u_j$
\begin{align}
{\tilde\psi}_{j}(x,u_j) &= \underset{\{\yb \in \Ycal:y_j= u_j\}}{\mathbf{\maximize}} \sum_e \psi_e(x,\yb_e). \label{eg_global} 
\end{align}
One readily sees (\ref{eg_global})  as a variant of the inference problem (\ref{eq_inference}), with similar solution techniques. 
The maximization operation fixes the labeling of the node $y_j=u_j$ and queries the optimal configuration for the remaining part of output graph.
In message-passing algorithms, only slight modification is needed to make sure that 
only the messages consistent with the microlabel restriction are considered. To obtain the vector $\tilde\Psi(x) = ( \tilde\psi_{j}(x,u_j))_{j,u_j}$ the same inference 
is repeated for each target-microlabel pair $(j,u_j)$, hence it has quadratic time complexity in the number of edges in the output graph.


Given the max-marginals of the base models, the Average of Max-Marginals (AMM) ensemble is constructed as follows.
Let $\mathcal{G}=\{G^{(1)},\cdots,G^{(T)}\}$ be a set of  output graphs, and let $\sets{\tilde\Psi^{(1)}(x),\cdots,\tilde\Psi^{(T)}(x)}$ be the max-marginal vectors of the base classifiers trained on the output graphs.
The ensemble prediction for each target is obtained by averaging the max-marginals  of the base models and choosing the maximizing microlabel for the node:
\begin{align*}
F^{\text{\tiny AMM}}_j(x) = \underset{u_j \in \Ycal_j}{\argmax} \frac{1}{|T|} \sum_{t=1}^{T}{\tilde\psi}_{j,u_j}^{(t)}(x),
\end{align*}
and the predicted multilabel is composed from the predicted  microlabels
\begin{align*}
F^{\text{\tiny AMM}}(x) = \left(F^{\text{\tiny AMM}}_j(x)\right)_{j \in V}.
\end{align*}

%

In principle, AMM ensemble can give different predictions compared to MVE, since the
most frequent label may not be the ensemble prediction if it has lower average max-marginal score.

\subsection{Maximum Average Marginals aggregation (MAM)}
\label{sec_ensemble3}

%
%

The next model, the Maximum of Average Marginals (MAM) ensemble, first collects the local compatibility scores $\Psi_E^{(t)}(x)$ from individual base learners, 
averages them and finally performs inference on the global consensus graph with averaged edge potentials. The model is defined as 
\begin{align*}
F^{\text{\tiny MAM}}(x) &= \underset{\yb \in \Ycal }{\argmax}\,\sum_{e\in E_t}\frac{1}{T}\sum_{t=1}^{T}\ \psi^{(t)}_e(x,\yb_e)
= \underset{\yb \in \Ycal}{\argmax}\, \frac{1}{T} \sum_{t=1}^T \sum_e \ip{\wb_e^{(t)}}{\varphi_e(x,\yb_e)}. 
\end{align*}

With the factorized dual representation, this ensemble scheme can be implemented simply and efficiently in terms of marginal dual variables and the associated kernels.
Using the Lemma (\ref{lemma_equ}) the above can be equivalently expressed as
\begin{align*}
F^{\text{\tiny MAM}}(x) 
 & =\underset{\yb \in \Ycal}{\argmax}\, \frac{1}{T} \sum_{t=1}^T \sum_{i,e,\ub_e} \mu^{(t)}(i,e,\ub_e) \cdot H_e(i,\ub_e;x,\yb_e)\nonumber \\
 & =\underset{\yb \in \Ycal}{\argmax}\,   \sum_{i,e,\ub_e}  \bar\mu(i,e,\ub_e) H_e(i,\ub_e;x,\yb_e), 
\end{align*}
where we denote by $ \bar\mu(i,e,\ub_e) =  \frac{1}{T} \sum_{t=1}^T \mu^{(t)}(i,e,\ub_e)$ the marginal dual variable averaged over the ensemble.
We note that $\mu^{(t)}$ is originally defined on edge set $E^{(t)}$,
$\mu^{(t)}$ from different random graph are not mutually consistent.
In practice, we first construct a consensus graph $\tilde{G}=(\tilde{E}, V)$ by pooling edge sets $E^{(t)}$, 
then complete $\mu^{(t)}$ on $\tilde{E}$ where missing components are computed via local consistency constraints.
Thus, the ensemble prediction can be computed in marginal dual form without explicit access to input features, and the only input needed from the 
different base models are the values of the marginal dual variables.

\subsection{The MAM Ensemble Analysis}

Here, we present theoretical analysis of the improvement of the MAM ensemble over the mean of the base classifiers. The analysis follows the spirit of the single-label ensemble  analysis by
\citet{brown2010good}, generalizing it to multilabel MAM ensemble. 

Assume there is a collection of $T$ individual base learners, indexed by $t\in\{1,\cdots,T\}$, that output compatibility scores $\psi^{(t)}_e(x,\ub_e)$ for all $t\in\{1,\dots,T\}$, $e \in E^{(t)}$, and $\ub_e \in \Ycal_e$.
For the purposes of this analysis, we express the compatibility scores in terms of the nodes (microlabels) instead of the edges and their labelings. We denote by 
$$\psi_j(x,y_j) =   \sum_{\substack{e = (j,j'),\\e\in N(j)}}  \ind{y_j =  u_j }\frac{1}{2}\psi_e(x,\ub_e)$$ 
the sum of compatibility scores of the set of edges $N(j)$ incident to node $j$ with  consistent labeling $\yb_e = (y_j,y_{j'}), y_j = u_j$.
Then, the compatibility score for the input and the multilabel in (\ref{compatibility_score}) can be alternatively expressed as
$$\psi(x,\yb) = \sum_{e \in E} \psi_e(x,\yb_e) = \sum_{j\in V} \psi_j(x,y_j).$$
%
%
The compatibility score from MAM ensemble can be similarly represented in terms of the nodes by
\begin{align*}
	\psi^{\text{\tiny MAM}}(x,\yb) =  \frac{1}{T}\sum_t \psi^{(t)}(x,\yb) =  \sum_{e \in E} \bar\psi_e(x,\yb_e) 
	= \sum_{j\in V} \bar\psi_j(x,y_j),
\end{align*}
where we have denoted $\bar\psi_j(x,y_j) = \frac{1}{T}\sum_t \psi^{(t)}_j(x,y_j)$ and $ \bar\psi_e(x,\yb_e) = \frac{1}{T}\sum_t  \psi^{(t)}_e(x,\yb_e)$.

Assume now the ground truth, the optimal compatibility score of an example and multilabel pair $(x,\yb)$,  is given by  $\psi^*(x,\yb)=\sum_{j\in V}\psi^*_j(x,y_j)$. We study the reconstruction error of the compatibility score distribution, given by the squared distance of
the estimated score distributions from the ensemble and the ground truth.
The reconstruction error of the MAM ensemble can be expressed as
\begin{align*}
	\Delta^R_{\text{\tiny MAM}}(x,\yb) = \left(\psi^*(x,\yb)-\psi^{\text{\tiny MAM}}(x,\yb)\right)^2,
\end{align*}
and the average reconstruction error of the base learners can be expressed as
\begin{align*}
	\Delta^R_I(x,\yb) = \frac{1}{T}\sum_{t}\left(\psi^*(x,\yb)-\psi^{(t)}(x,\yb)\right)^2.
\end{align*}

We denote by $\Psi_j(x,y_j)$  a random variable of the compatibility scores obtained by the base learners and $\{\psi^{(1)}_j(x,y_j),\cdots,\psi^{(T)}_j(x,y_j)\}$ as a sample from its distribution.  We have the following result:

\begin{theorem}
	\label{t1}
The reconstruction error of compatibility score distribution given by MAM ensemble $\Delta^R_{\text{\tiny MAM}}(x,\yb)$ is guaranteed to be no greater than the average reconstruction error given by individual base learners $\Delta^R_I(x,\yb)$. 

In addition, the gap can be estimated as 
\begin{align*}
	\Delta^R_I(x,\yb) - \Delta^R_{\text{\tiny MAM}}(x,\yb) &= \var(\sum_{j\in V}\Psi_j(x,y_j)) \ge 0.
\end{align*}
The variance can be further expanded as 
\begin{align*}
	\var(\sum_{j\in V}\Psi_j(x,y_j)) & = \underbrace{\sum_{\substack{j\in V\\\,}}\var(\Psi_j(x,y_j))}_{diversity} + \underbrace{\sum_{\substack{p,q\in V,\\p\ne q}}\cov(\Psi_{p}(x,y_p),\Psi_{q}(x,y_q))}_{coherence}.
\end{align*}
\end{theorem}

\begin{proof}
By expanding and simplifying the squares we get
\begin{align*}
	\Delta^R_I(x,\yb) - &\Delta^R_{\text{\tiny MAM}}(x,\yb) 
	= \frac{1}{T}\sum_{t}\left(\psi^*(x,\yb)-\psi^{(t)}(x,\yb)\right)^2 - \left(\psi^*(x,\yb) - \psi^{\text{\tiny MAM}}(x,\yb)\right)^2\\
	&= \frac{1}{T}\sum_{t}\left(\sum_{j\in V}\psi^*_j(x,y_j)-\sum_{j\in V}\psi^{(t)}_j(x,y_j)\right)^2 - \left(\sum_{j\in V}\psi^*_j(x,y_j) - \sum_{j\in V}\frac{1}{T}\sum_t \psi^{(t)}_j(x,y_j)\right)^2\\
	&= \frac{1}{T}\sum_{t}\left(\sum_{j\in V}\psi^{(t)}_j(x,y_j)\right)^2 - \left(\frac{1}{T}\sum_t \sum_{j\in V}\psi^{(t)}_j(x,y_j)\right)^2\\
	& = \var(\sum_{j\in V}\Psi_j(x,y_j))\\
	& \ge 0.
\end{align*}
The expression of variance can be further expanded as
\begin{align*}
	\var(\sum_{j\in V}\Psi_j(x,y_j)) & = \sum_{p,q\in V}\cov(\Psi_p(x,y_p),\Psi_q(x,y_q))\\
	&= {\sum_{j\in V}\var(\Psi_j(x,y_j))}
	+ {\sum_{\substack{p,q\in V,\\p\ne q}}\cov(\Psi_{p}(x,y_p),\Psi_{q}(x,y_q))}.
\end{align*}
\end{proof}
The Theorem~\ref{t1} states that the reconstruction error from MAM ensemble is guaranteed to be less than or equal to the average reconstruction error from the individuals.
In particular, the improvement can be further addressed by two terms, namely {\em diversity} and {\em coherence}.
The classifier diversity measures the variance of predictions from base learners independently on each single labels.
It has been previously studied in single-label classifier ensemble context by \citet{Krogh95}. 
The diversity term prefers the variability of individuals that learn from different perspectives. It is a well known factor to improve the ensemble performance.
The coherence term, that is specific to the multilabel classifiers, indicates that
the more the microlabel predictions vary together,
the greater advantage multilabel ensemble gets over the base learners.
This supports our intuitive understanding that microlabel correlations are keys to successful multilabel learning.

%
%

\section{Experiments}
\label{sec_experiments}

\subsection{Datasets}
We experiment on a  collection of ten multilabel datasets from different domains, including chemical, biological, and text classification problems. 
The {\em NCI60} dataset contains $4547$ drug candidates with their cancer inhibition potentials in $60$ cell line targets. 
The {\em Fingerprint} dataset links $490$ molecular mass spectra together to $286$ molecular substructures used as prediction targets.
Four text classification datasets\footnote{Available at http://mulan.sourceforge.net/datasets.html} are also used in our experiment. In addition, 
two artificial {\em Circle} dataset are generated according to \citep{wei2012corrlog} with different amount of labels.
An overview of the datasets is shown in Table~\ref{experiment_table_1},
where {\em cardinality} is the average number of positive microlabels in the examples, defined as
\begin{align*}
	cardinality =\frac{1}{m}\sum_{i=1}^m | \{ j|\yb_{ij} = 1\}|,
\end{align*}
and {\em density} is the average number of labels of examples divided by the size of label space as
\begin{align*}
	density= cardinality/k.
\end{align*}

\begin{table}[t]
	\small
\caption{Statistics of multilabel datasets used in our experiments. For {\em NCI60} and {\em Fingerprint} dataset where there is no explicit feature representation, the rows of kernel matrix is assumed as feature vector.}
\label{experiment_table_1}
\begin{center}
\begin{sc}
\begin{tabular}{|l|c|c|c|c|c|} \hline
	\multirow{2}{*}{\textbf{Dataset}} 
	& \multicolumn{5}{c|}{\textbf{Statistics}} \\ \cline{2-6}
	& \textbf{Instances} & \textbf{Labels} & \textbf{Features} & \textbf{Cardinality} & \textbf{Density} \\ \hline
Emotions & $593$ & $6$ & $72$ & $1.87$ & $0.31$ \\ \hline
Yeast & $2417$ & $14$ & $103$ & $4.24$ & $0.30$ \\ \hline
Scene & $2407$ & $6$ & $294$ & $1.07$ & $0.18$ \\ \hline
Enron & $1702$ & $53$ & $1001$ & $3.36$ & $0.06$ \\ \hline
Cal500 & $502$ & $174$ & $68$ & $26.04$ & $0.15$ \\ \hline
Fingerprint & $490$ & $286$ & $490$ & $49.10$ & $0.17$ \\ \hline
NCI60 & $4547$ & $60$ & $4547$ & $11.05$ & $0.18$ \\ \hline
Medical & $978$ & $45$ & $1449$ & $1.14$ & $0.03$ \\ \hline
Circle10 & $1000$ & $10$ & $3$ & $8.54$ & $0.85$ \\ \hline
CIrcle50 & $1000$ & $50$ & $3$ & $35.63$ & $0.71$ \\ \hline
\end{tabular}
\end{sc}
\end{center}
\end{table}

We calculate linear kernel on datasets where examples are described by feature vectors.
For text classification datasets, we first compute TF-IDF weighted features.
For {\em Fingerprint} datasets we compute quadratic kernel over the 'bag' of mass/charge peak intensities in the MS/MS spectra.
On this dataset, as feature vectors for non-kernelized methods the rows of the training kernel matrix are used, due to the intractability of using the 
explicit features.

\subsection{Compared Classification Methods}
For comparison, we choose the following established classification methods form different perspectives towards multilabel classification,  accounting 
for single-label and multilabel, as well as ensemble and standalone methods:

\begin{itemize}
\item Support Vector Machine (SVM) is used as the single-label non-ensemble baseline classification model.
In practice, we train a collection of SVMs, one for each microlabel.

\item{Bagging} \citep{Breiman96} is used as the benchmark single-label ensemble method. 
In practice, we randomly select $40\%$ of the data as input to SVM to get a weak hypothesis, and repeat the process until we collect an ensemble of $60$ weak hypotheses.

\item{MMCRF} \citep{rousu2007} is used both as a standalone multilabel classifier and the base classifier in the ensembles.
Individual MMCRF models are trained with random tree as output graph structures.

\item{Multi-task feature learning} (MTL), proposed in \citep{Argyriou07multi-taskfeature}, is used as another multilabel  benchmark.

\item{AdaBoostMH} is a multilabel variant of AdaBoost developed in \citep{SchSin00}. In our study, we use real-valued decision tree with at most $100$ leaves as base 
learner of AdaBoostMH. We successively generate an ensemble of 100 weak hypothesises.

\end{itemize}

\subsection{Obtaining Random Output Graphs}\label{sec_outputgraphs}

Output graphs for the graph labeling classifiers are generated by first drawing a random $k \times k$ matrix  with non-negative edge weights and then extracting a maximum weight spanning tree
out of the matrix.  The spanning tree connects all targets so that the complex microlabel dependencies can be learned. Also, the tree structure facilitates efficient inference.

\subsection{Parameter Selection and Evaluation Measures}

We first sample $10\%$ data uniform at random from each experimental dataset for parameter selection.
Both SVM and MMCRF base models have margin softness parameter $C$, which potentially need to be tuned.
We tested parameter $C$ from a set $\{0.01,0.1,0.5,1,5,10\}$ based on tuning data for both SVM and base learner MMCRF, then keep the best ones for the following validation step.
We also perform extensive selection on $\gamma$ parameters in MTL model in the same range as margin softness parameters. 

Because most of the multilabel datasets are highly biased with regards to multilabel density, we use the following {\em stratified $5$-fold cross validation} scheme in the experiments reported, such that we group examples in equivalent classes based on the number of positive labels they have. 
Each equivalent class is then randomly split into five local folds, after that the local folds are merged to create five global folds.
The proposed procedure ensures that also the smaller classes have representations in all folds. 

To quantitatively evaluate the performance of different classifiers, we adopt several performance measures.
We report {\em multilabel accuracy} which counts the proportion of multilabel predictions that have all of the microlabels being correct,
{\em microlabel accuracy} as the proportion of microlabel being correct,
and {\em microlabel $F_1$ score} that is the harmonic mean of microlabel precision and recall $F_1 = 2\cdot\frac{Pre\times Rec}{Pre + Rce}$.

\subsection{Comparison of Different Ensemble Approaches}

\begin{figure}[t]
\begin{center}
\includegraphics[page={1},width=1 \columnwidth]{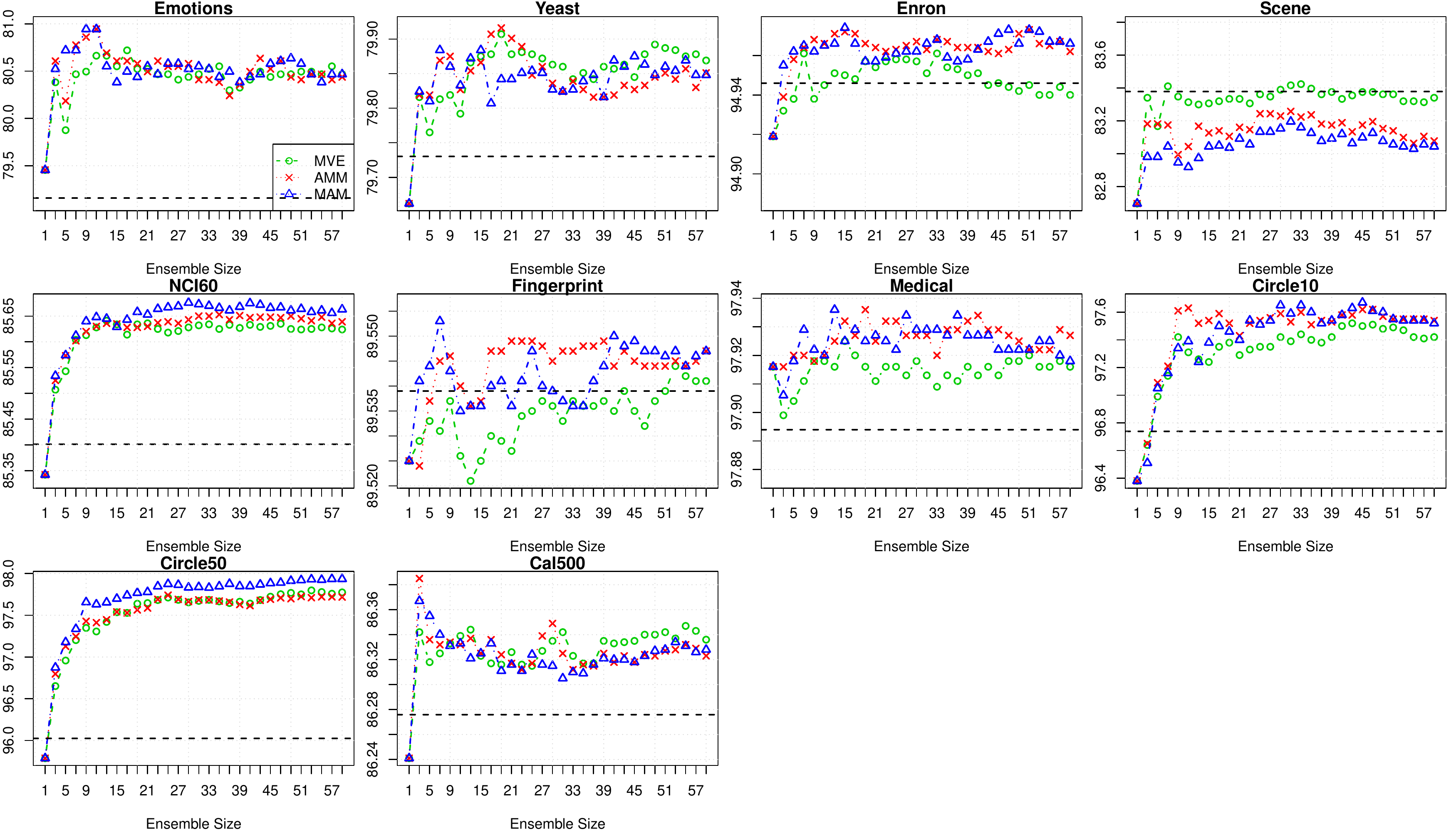}
\caption{Ensemble learning curve (microlabel accuracy) plotted as the size of ensemble. Average performance of base learner with random tree as output graph structure is denoted as horizontal dash line.}
\label{ensemblemethodeffect}
\end{center}
\end{figure}

Figure~\ref{ensemblemethodeffect} depicts the ensemble learning curves in varying datasets with respect to microlabel accuracy.
In general, there is a clear trend of improving microlabel accuracy for random tree based ensemble approaches as more individual base models are combined. 
We observe the similar trends in multilabel accuracy and microlabel $F_1$ space (see supplementary material for plots).
We also notice that most of the learning curves converge even with a small ensemble size.

All three proposed ensemble learners (MVE, AMM, MAM) outperform their base learner MMCRF (horizontal dash lines) with significant margins in almost all datasets,
the {\em Scene} being the only exception.
AMM and MAM outperform MVE in all datasets except for {\em Scene} and {\em Cal500}.
Furthermore, MAM approach surpasses AMM in nine out of ten datasets. 
Consequently, we choose MAM for the further studies described in the following section.

\subsection{Multilabel Prediction Performance}

\begin{table}
	\small
	\caption{Prediction performance of each algorithm in terms of microlabel accuracy, multilabel accuracy, and microlabel $F_1$ score ('$-$' denotes no positive predictions). {\em $@$Top2} counts how many times the algorithm achieves at least the second best.}
	\label{prediction_performance}
	\begin{center}
	\begin{sc}
	\begin{tabular}{|c|p{1.6cm}|p{1.6cm}|p{1.8cm}|p{1.6cm}|p{1.6cm}|p{1.6cm}|} \hline
		\multirow{2}{*}{\textbf{Dataset}} 
		& \multicolumn{6}{c|}{\textbf{Microlabel Accuracy}} \\ \cline{2-7} 
	    & Svm & Bagging & AdaBoost & Mtl & Mmcrf & Mam\\ \cline{1-7}
		Emotions & {77.3}$\pm${1.9} & {74.1}$\pm${1.8} & {76.8}$\pm${1.6} & \em{79.8}$\pm${1.8} & {79.2}$\pm${0.9} & \bf{80.5}$\pm${1.4} \\ \hline
		Yeast & \bf{80.0}$\pm${0.6} & {78.4}$\pm${0.7} & {74.8}$\pm${0.3} & {79.3}$\pm${0.2} & {79.7}$\pm${0.3} & \em{79.9}$\pm${0.4} \\ \hline
		Scene & \bf{90.2}$\pm${0.3} & {87.8}$\pm${0.8} & {84.3}$\pm${0.4} & \em{88.4}$\pm${0.6} & {83.4}$\pm${0.2} & {83.0}$\pm${0.2} \\ \hline
		Enron & {93.6}$\pm${0.2} & {93.7}$\pm${0.1} & {86.2}$\pm${0.2} & {93.5}$\pm${0.1} & \em{94.9}$\pm${0.1} & \bf{95.0}$\pm${0.2} \\ \hline
		Cal500 & \bf{86.3}$\pm${0.3} & {86.0}$\pm${0.2} & {74.9}$\pm${0.4} & {86.2}$\pm${0.2} & \bf{86.3}$\pm${0.2} & \bf{86.3}$\pm${0.3} \\ \hline
		Fingerprint & \bf{89.7}$\pm${0.2} & {85.0}$\pm${0.7} & {84.1}$\pm${0.5} & {82.7}$\pm${0.3} & \em{89.5}$\pm${0.3} & \em{89.5}$\pm${0.8} \\ \hline
		NCI60 & {84.7}$\pm${0.7} & {79.5}$\pm${0.8} & {79.3}$\pm${1.0} & {84.0}$\pm${1.1} & \em{85.4}$\pm${0.9} & \bf{85.7}$\pm${0.7} \\ \hline
		Medical & {97.4}$\pm${0.1} & {97.4}$\pm${0.1} & {91.4}$\pm${0.3} & {97.4}$\pm${0.1} & \bf{97.9}$\pm${0.1} & \bf{97.9}$\pm${0.1} \\ \hline
		Circle10 & {94.8}$\pm${0.9} & {92.9}$\pm${0.9} & \bf{98.0}$\pm${0.4} & {93.7}$\pm${1.4} & {96.7}$\pm${0.7} & \em{97.5}$\pm${0.3} \\ \hline
		Circle50 & {94.1}$\pm${0.3} & {91.7}$\pm${0.3} & \em{96.6}$\pm${0.2} & {93.8}$\pm${0.7} & {96.0}$\pm${0.1} & \bf{97.9}$\pm${0.2} \\ \hline
		$@Top2$ & {4} & {0} & {2} & {2} & \em{5} & \bf{9} \\ \hline
		\hline
		\multirow{2}{*}{\textbf{Dataset}} 
		& \multicolumn{6}{c|}{\textbf{Multilabel Accuracy}} \\ \cline{2-7} 
	    & Svm & Bagging & AdaBoost & Mtl & Mmcrf & Mam\\ \cline{1-7}
		Emotions & {21.2}$\pm${3.4} & {20.9}$\pm${2.6} & {23.8}$\pm${2.3} & {25.5}$\pm${3.5} & \em{26.5}$\pm${3.1} & \bf{30.4}$\pm${4.2} \\ \hline
		Yeast & \em{14.0}$\pm${1.8} & {13.1}$\pm${1.2} & {7.5}$\pm${1.3} & {11.3}$\pm${2.8} & {13.8}$\pm${1.5} & \bf{14.0}$\pm${0.6} \\ \hline
		Scene & \bf{52.8}$\pm${1.0} & \em{46.5}$\pm${2.5} & {34.7}$\pm${1.8} & {44.8}$\pm${3.0} & {12.6}$\pm${0.7} & {5.4}$\pm${0.5} \\ \hline
		Enron & {0.4}$\pm${0.1} & {0.1}$\pm${0.2} & {0.0}$\pm${0.0} & {0.4}$\pm${0.3} & \em{11.7}$\pm${1.2} & \bf{12.1}$\pm${1.0} \\ \hline
		Cal500 & {0.0}$\pm${0.0} & {0.0}$\pm${0.0} & {0.0}$\pm${0.0} & {0.0}$\pm${0.0} & {0.0}$\pm${0.0} & {0.0}$\pm${0.0} \\ \hline
		Fingerprint & \bf{1.0}$\pm${1.0} & {0.0}$\pm${0.0} & {0.0}$\pm${0.0} & {0.0}$\pm${0.0} & \em{0.4}$\pm${0.9} & \em{0.4}$\pm${0.5} \\ \hline
		NCI60 & \em{43.1}$\pm${1.3} & {21.1}$\pm${1.3} & {2.5}$\pm${0.6} & \bf{47.0}$\pm${1.4} & {36.9}$\pm${0.8} & {40.0}$\pm${1.0} \\ \hline
		Medical & {8.2}$\pm${2.3} & {8.2}$\pm${1.6} & {5.1}$\pm${1.0} & {8.2}$\pm${1.2} & \em{35.9}$\pm${2.1} & \bf{36.9}$\pm${4.6} \\ \hline
		Circle10 & {69.1}$\pm${4.0} & {64.8}$\pm${3.2} & \bf{86.0}$\pm${2.0} & {66.8}$\pm${3.4} & {75.2}$\pm${5.6} & \em{82.3}$\pm${2.2} \\ \hline
		Circle50 & {29.7}$\pm${2.5} & {21.7}$\pm${2.6} & {28.9}$\pm${3.6} & {27.7}$\pm${3.4} & \em{30.8}$\pm${1.9} & \bf{53.8}$\pm${2.2} \\ \hline
		$@Top2$ & {5} & {2} & {2} & {2} & \em{6} & \bf{8} \\ \hline
		\hline
		\multirow{2}{*}{\textbf{Dataset}} 
		& \multicolumn{6}{c|}{\textbf{Microlabel $F_1$ Score}} \\ \cline{2-7} 
	    & Svm & Bagging & AdaBoost & Mtl & Mmcrf & Mam\\ \cline{1-7} 
		Emotions & {57.1}$\pm${4.4} & {61.5}$\pm${3.1} & \em{66.2}$\pm${2.9} & {64.6}$\pm${3.0} & {64.6}$\pm${1.2} & \bf{66.3}$\pm${2.3} \\ \hline
		Yeast & {62.6}$\pm${1.2} & \bf{65.5}$\pm${1.3} & \em{63.5}$\pm${0.6} & {60.2}$\pm${0.5} & {62.4}$\pm${0.7} & {62.4}$\pm${0.6} \\ \hline
		Scene & \em{68.3}$\pm${0.9} & \bf{69.9}$\pm${1.9} & {64.8}$\pm${0.8} & {61.5}$\pm${2.4} & {23.7}$\pm${1.2} & {11.6}$\pm${0.9} \\ \hline
		Enron & {29.4}$\pm${1.0} & {38.8}$\pm${1.5} & {42.3}$\pm${1.1} & {-} & \bf{53.8}$\pm${1.3} & \em{53.7}$\pm${0.7} \\ \hline
		Cal500 & {31.4}$\pm${0.8} & \em{40.1}$\pm${0.3} & \bf{44.3}$\pm${0.5} & {28.6}$\pm${0.6} & {32.7}$\pm${0.9} & {32.3}$\pm${0.9} \\ \hline
		Fingerprint & \bf{66.3}$\pm${0.8} & {64.4}$\pm${1.9} & {62.8}$\pm${1.6} & {0.4}$\pm${0.4} & \em{65.0}$\pm${1.4} & \em{65.0}$\pm${2.1} \\ \hline
		NCI60 & {45.9}$\pm${1.9} & \bf{53.9}$\pm${1.3} & {32.9}$\pm${2.0} & {32.9}$\pm${0.9} & {46.7}$\pm${2.8} & \em{47.1}$\pm${2.9} \\ \hline
		Medical & {-} & {-} & {33.7}$\pm${1.1} & {-} & \em{49.5}$\pm${3.5} & \bf{50.3}$\pm${3.5} \\ \hline
		Circle10 & {97.0}$\pm${0.5} & {96.0}$\pm${0.5} & \bf{98.8}$\pm${0.2} & {96.4}$\pm${0.9} & {98.1}$\pm${0.4} & \em{98.6}$\pm${0.2} \\ \hline
		Circle50 & {96.0}$\pm${0.3} & {94.5}$\pm${0.2} & \em{97.6}$\pm${0.1} & {95.7}$\pm${0.5} & {97.2}$\pm${0.1} & \bf{98.6}$\pm${0.1} \\ \hline
		$@Top2$	 & {2} & {4} & \em{5} & {0} & {3} & \bf{7} \\ \hline
	\end{tabular}
	\end{sc}
	\end{center}
\end{table}

We examine whether our proposed ensemble model (MAM) can boost the prediction performance in multilabel classification problems. 
Therefore, we compare our model with other advanced methods including both single-label and multilabel classifiers, both standalone and ensemble frameworks. 
Table~\ref{prediction_performance} shows the performance of difference methods in terms of microlabel accuracy, multilabel accuracy and microlabel $F_1$ score, where 
the best performance in each dataset is emphasised in {\bf boldface} and the second best is in {\em italics}.
We also count how many times each algorithm achieves at least the second best performance.
The total count is shown as {\em '@Top2'}.

We observe from Table~\ref{prediction_performance} that MAM outperforms both standalone and ensemble competitors in all three measurements.
In particular, it is ranked nine times as top 2 methods in microlabel accuracy, eight times in multilabel accuracy, and seven times in microlabel $F_1$ score.
The only datasets where MAM is consistently outside the top 2 is the {\em Scene} dataset. The dataset is practically a single-label multiclass dataset, with very few
examples with more than one positive microlabel.
The graph-based approaches MMCRF and MAM do not seem to be able to cope with the extreme label sparsity.
However, on this dataset the single target classifiers SVM and Bagging outperform all compared multilabel classifiers.

In these experiments, MMCRF also performs robustly, being in top 2 on half of the datasets with respect to microlabel and multilabel accuracy, however,
quite consistently trailing to MAM, often with a noticeable margin.

We also notice that the standalone single target classifier SVM is competitive against most multilabel methods,
placing in top 2 more often than Bagging, AdaBoost and MTL with respect to microlabel and microlabel accuracy.

Overall, the results indicate that ensemble by MAM is a robust and competitive alternatives for multilabel classification.

\section{Conclusions}
 \label{sec_conclusions}
 
In this paper we have put forward new methods for multilabel classification, relying on ensemble learning on random output graphs. 
In our experiments, models thus created have favourable predictive performances on a heterogeneous collection of multilabel datasets, compared to several established methods. 
The theoretical analysis of the MAM ensemble highlights the covariance of the compatibility scores between the inputs and microlabels learned by the base learners
as the quantity explaining the advantage of the ensemble prediction over the base learners.
Our results indicate that structured output prediction methods can be successfully applied to problems where no prior known output structure exists, and thus widen the applicability of the 
structured output prediction.

We leave it as an open problem to analyze the generalization error of this type classifiers.
We also plan to link diversity term to model performance through empirical evaluations.

\acks{
The work was financially supported by Helsinki Doctoral Programme in Computer Science (Hecse), 
Academy of Finland grant 118653 (ALGODAN), 
IST Programme of the European Community under the PASCAL2 Network of Excellence, ICT-2007-216886. 
This publication only reflects the authors' views.
}


\vskip 0.2in
\bibliography{su65}

\begin{thebibliography}{14}
\providecommand{\natexlab}[1]{#1}
\providecommand{\url}[1]{\texttt{#1}}
\expandafter\ifx\csname urlstyle\endcsname\relax
  \providecommand{\doi}[1]{doi: #1}\else
  \providecommand{\doi}{doi: \begingroup \urlstyle{rm}\Url}\fi

\bibitem[Argyriou et~al.(2007)Argyriou, Evgeniou, and
  Pontil]{Argyriou07multi-taskfeature}
Andreas Argyriou, Theodoros Evgeniou, and Massimiliano Pontil.
\newblock Multi-task feature learning.
\newblock In \emph{Advances in Neural Information Processing Systems 19}. MIT
  Press, 2007.

\bibitem[Bian et~al.(2012)Bian, Xie, and Tao]{wei2012corrlog}
Wei Bian, Bo~Xie, and Dacheng Tao.
\newblock Corrlog: Correlated logistic models for joint prediction of multiple
  labels.
\newblock In \emph{Proceedings of the Fifteenth International Conference on
  Artificial Intelligence and Statistics (AISTATS-12)}, volume~22, pages
  109--117, 2012.

\bibitem[Breiman(1996)]{Breiman96}
Leo Breiman.
\newblock Bagging predictors.
\newblock \emph{Machine Learning}, 24:\penalty0 123--140, 1996.

\bibitem[Brown and Kuncheva(2010)]{brown2010good}
Gavin Brown and Ludmila~I Kuncheva.
\newblock ÒgoodÓ and ÒbadÓ diversity in majority vote ensembles.
\newblock In \emph{Multiple Classifier Systems}, pages 124--133. Springer,
  2010.

\bibitem[Esuli et~al.(2008)Esuli, Fagni, and Sebastiani]{Esu08}
A.~Esuli, T.~Fagni, and F.~Sebastiani.
\newblock Boosting multi-label hierarchical text categorization.
\newblock \emph{Information Retrieval}, 11\penalty0 (4):\penalty0 287--313,
  2008.

\bibitem[Krogh and Vedelsby(1995)]{Krogh95}
Anders Krogh and Jesper Vedelsby.
\newblock Neural network ensembles, cross validation, and active learning.
\newblock In \emph{Advances in Neural Information Processing Systems}, pages
  231--238. MIT Press, 1995.

\bibitem[Rousu et~al.(2006)Rousu, Saunders, Szedmak, and
  Shawe-Taylor]{rousu2006kbl}
J.~Rousu, C.~Saunders, S.~Szedmak, and J.~Shawe-Taylor.
\newblock {Kernel-Based Learning of Hierarchical Multilabel Classification
  Models}.
\newblock \emph{The Journal of Machine Learning Research}, 7:\penalty0
  1601--1626, 2006.

\bibitem[Rousu et~al.(2007)Rousu, Saunders, Szedmak, and
  Shawe-Taylor]{rousu2007}
J.~Rousu, C.~Saunders, S.~Szedmak, and J.~Shawe-Taylor.
\newblock {Efficient algorithms for max-margin structured classification}.
\newblock \emph{Predicting Structured Data}, pages 105--129, 2007.

\bibitem[Schapire and Singer(2000)]{SchSin00}
Robert~E. Schapire and Yoram Singer.
\newblock Boostexter: {A} boosting-based system for text categorization.
\newblock \emph{Machine Learning}, 39\penalty0 (2/3):\penalty0 135 -- 168,
  2000.

\bibitem[Su and Rousu(2011)]{su2011multi}
H.~Su and J.~Rousu.
\newblock Multi-task drug bioactivity classification with graph labeling
  ensembles.
\newblock \emph{Pattern Recognition in Bioinformatics}, pages 157--167, 2011.

\bibitem[Taskar et~al.(2003)Taskar, Guestrin, and Koller]{TGK:nips03}
B.~Taskar, C.~Guestrin, and D.~Koller.
\newblock Max-margin markov networks.
\newblock In \emph{Neural Information Processing Systems}, 2003.

\bibitem[Tsochantaridis et~al.(2004)Tsochantaridis, Hofmann, Joachims, and
  Altun]{THJA:icml04}
I.~Tsochantaridis, T.~Hofmann, T.~Joachims, and Y.~Altun.
\newblock Support vector machine learning for interdependent and structured
  output spaces.
\newblock In \emph{ICML'04}, pages 823--830, 2004.

\bibitem[Wainwright et~al.(2005)Wainwright, Jaakkola, and
  Willsky]{wainwright2005map}
M.J. Wainwright, T.S. Jaakkola, and A.S. Willsky.
\newblock {MAP estimation via agreement on trees: message-passing and linear
  programming}.
\newblock \emph{IEEE Transactions on Information Theory}, 51\penalty0
  (11):\penalty0 3697--3717, 2005.

\bibitem[Yan et~al.(2007)Yan, Tesic, and Smith]{yan2007model}
R.~Yan, J.~Tesic, and J.R. Smith.
\newblock Model-shared subspace boosting for multi-label classification.
\newblock In \emph{Proceedings of the 13th ACM SIGKDD international conference
  on Knowledge discovery and data mining}, pages 834--843. ACM, 2007.

\end{thebibliography}

\end{document}